\documentclass[a4paper]{article}

\usepackage{INTERSPEECH2021}

\title{The Role of Phonetic Units in Speech Emotion Recognition}
\name{Jiahong Yuan, Xingyu Cai, Renjie Zheng, Liang Huang, Kenneth Church}
\address{
  Baidu Research, USA}
\email{\{jiahongyuan,xingyucai,renjiezheng,lianghuang,kennethchurch\}@baidu.com}

\begin{document}

\maketitle
\begin{abstract}
We propose a method for emotion recognition through emotion-dependent speech recognition using Wav2vec 2.0. Our method achieved a significant improvement over most previously reported results on IEMOCAP, a benchmark emotion dataset. Different types of phonetic units are employed and compared in terms of accuracy and robustness of emotion recognition within and across datasets and languages. Models of phonemes, broad phonetic classes, and syllables all significantly outperform the utterance model, demonstrating that phonetic units are helpful and should be incorporated in speech emotion recognition. The best performance is from using broad phonetic classes. Further research is needed to investigate the optimal set of broad phonetic classes for the task of emotion recognition. Finally, we found that Wav2vec 2.0 can be fine-tuned to recognize coarser-grained or larger phonetic units than phonemes, such as broad phonetic classes and syllables.
\end{abstract}

\section{Introduction}
Speech emotion recognition is essentially a sequence-to-one classification problem whereas speech recognition is a sequence-to-sequence problem. In this paper we attempt to bridge these two problems. More specifically, we propose a method for emotion recognition through emotion-dependent speech recognition.

Speech emotion recognition has witnessed a stable advancement over the last two decades \cite{Ververidis&Kotropoulos,Ayadietal,Koolagudi&Rao,Anagnostopoulosetal,akccay2020speech}. Much of the earlier effort was concentrated on feature engineering. The emobase feature set in the widely used OpenSMILE toolkit \cite{Eybenetal}, for example, consists of 998 acoustic features for emotion recognition, including prosodic, spectral, as well as voice quality features. In recent years, more effort has been devoted to improving deep learning model architectures. Many studies in the literature are based on IEMOCAP \cite{busso2008iemocap}, a benchmark emotion dataset. At ICASSP 2020, half a dozen papers reported a cross-validation accuracy of 70\% or better on this dataset \cite{wangetal,liuetal,luetal,Pappagarietal,Yehetal}, based on acoustic data only. \cite{wangetal} achieved the highest accuracy of 73\% from employing a dual-sequence LSTM model.

A confounding factor for emotion recognition is the acoustic variability of different phonetic units such as phonemes. Figure 1 compares the spectra of two vowels (from the same speaker) in the same emotion. Clearly, their spectra are different in terms of the location of spectral peaks, which determines vowel quality. There are two approaches to overcome this problem. The predominant one is to make emotion features and models independent and more robust to phonetic variability. The other approach is to take consideration of phonetic variability by developing phonetically-aware features and models. This study adopts the second approach.

A relatively small number of studies of speech emotion recognition have investigated the effect of phonetic variability. \cite{Shahetal} incorporated articulatory information in emotion recognition. They observed that for the vowel /AE/ anger forces a larger opening of the jaw as opposed to sadness, while for the vowel /IY/ anger makes lips more protruded towards the outside. \cite{Bitouketal} computed statistics of Mel-Frequency Cepstral Coefficients over three phonetic classes (stressed vowels, unstressed vowels, and consonants) respectively for emotion recognition. They found that spectral features computed from consonant regions of the utterance contain more information about emotion than either stressed or unstressed vowel features. \cite{Schulleretal} investigated whether acoustic emotion recognition strongly depends on phonetic content. They demonstrated that phoneme-specific emotion models can lead to higher accuracies.  \cite{Dhamyal2020ThePB} used a self-attention based emotion classification model to understand the phonetic bases of emotions by discovering the most “attended” phonemes for each emotion. They found that the distribution of these attended-phonemes varies significantly between natural versus acted emotions.

To make use of the effect of phonetic variability for emotion recognition, we train emotion-dependent speech recognition models. For example, the vowel /AA/ in “happy” and “sad” are different phonemes and have different acoustic models. With emotion-dependent speech recognition models, emotion recognition can be done as a by-product of speech recognition. Similar efforts have been made in the literature. \cite{Vlasenko&Wendemuth} trained two sets of HMM/GMM acoustic models of phonemes for high-arousal and low-arousal/neutral emotions respectively, and determined the emotion of each phoneme in an utterance (generated from speech recognition) by applying and comparing the two models of the phoneme. Compared to \cite{Vlasenko&Wendemuth}, our approach is simpler. We combine speech recognition and emotion classification into one step of recognizing emotion-dependent phonetic units such as phonemes.

Emotion-dependent models may require more training data than emotion-independent models because the number of units/models is increased by multiple times (\textit{n} times where \textit{n} is the number of emotions). However, with the recent advancement of pretrained acoustic models, the amount of data needed for training speech recognition models is greatly reduced. For example, Wav2vec 2.0 \cite{baevski2020wav2vec} outperforms the previous state of the art on the Librispeech \cite{panayotov2015librispeech} test set in terms of word error rate while using 100 times less labeled data. By fine-tuning Wav2vec 2.0, we can train well-performing emotion-dependent models even with a small amount of data. 

We can also reduce the size of phonetic units in emotion-dependent models by using coarser-grained (e.g., broad phonetic classes) or larger (e.g., syllables) units than phonemes. Besides to increase the training data for each unit, we explore this idea in the study from three perspectives: 1. Can pretrained acoustic representations be fine-tuned to recognize coarser or larger phonetic units than phonemes? 2. What are the best phonetic units for emotion recognition? 3. Given that phonemes are language-specific while broad phonetic classes and syllables are language-general, are broad-phonetic-class or syllable models more robust than phoneme models for cross-lingual emotion recognition?

In the following sections we will first introduce the method of fine-tuning Wav2vec 2.0 for emotion recognition, followed by experiments on the English IEMOCAP dataset and three other datasets, in German, Arabic, and Mandarin Chinese respectively. Conclusions and discussion are made in the last section. 

\begin{figure}[t]
  \centering
  \includegraphics[width=\linewidth]{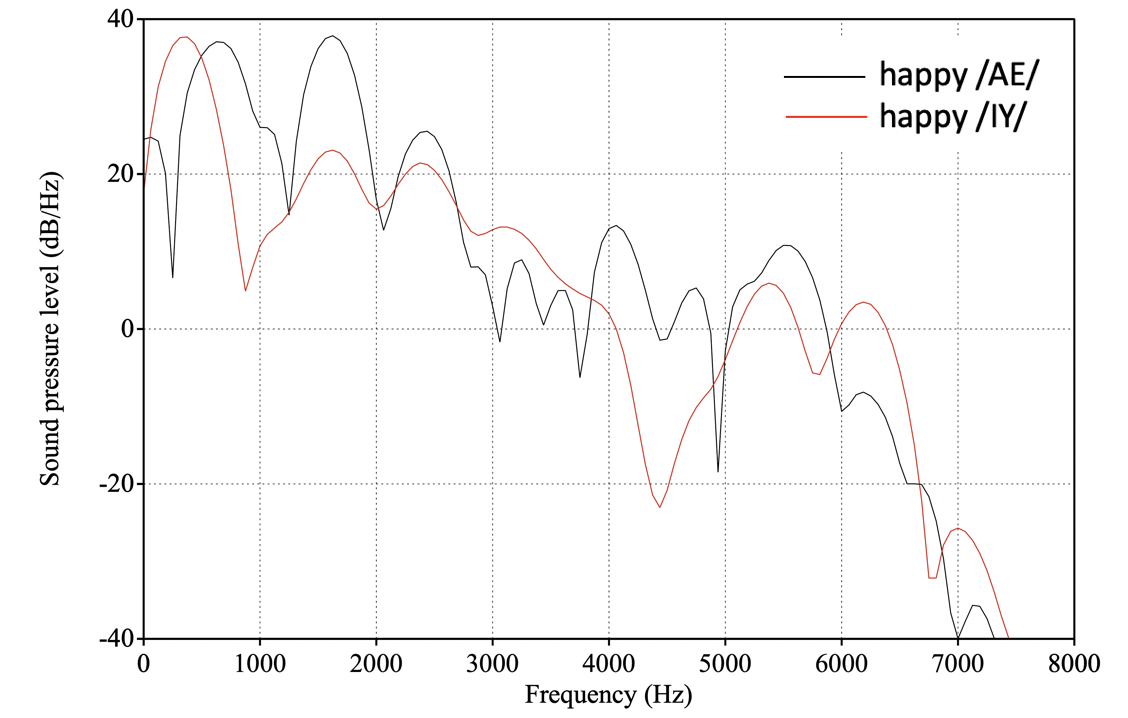}
  \caption{Spectra of happy /AE/ and /IY/.}
  \label{fig:Spectra of /AE/ and /IY/}
\end{figure}

\section{Fine-tuning Wav2vec2.0 for emotion recognition}

\subsection{The procedure}
Wav2vec 2.0 is a framework for self-supervised learning of speech representations. It consists of multiple convolution layers and self-attention layers, and pretrains on audio alone by masking the speech input in the latent space and solving a contrastive task defined over a quantization of the latent representations. Pre-trained wav2vec models can be fine-tuned for speech recognition with labeled data. 

\begin{figure}[t]
  \centering
  \includegraphics[width=\linewidth]{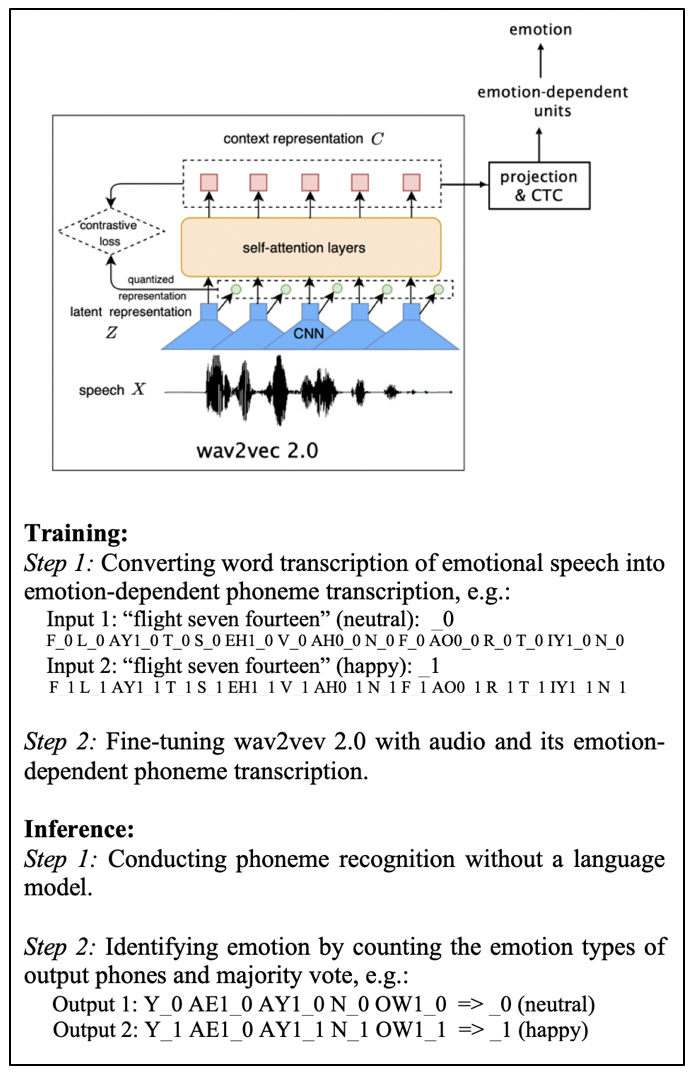}
  \caption{Fine-tuning wav2vec 2.0 for emotion recognition.}
  \label{fine-tuning}
\end{figure}

The procedure of fine-tuning Wav2vec 2.0 for emotion recognition is illustrated and described in Figure~\ref{fine-tuning}. The core idea is to use emotion-dependent units as labeled targets. A randomly initialized linear projection is added on top of the contextual representations of Wav2vec 2.0 to map the representations into emotion-dependent units (i.e., classes), and the entire model is optimized by minimizing the CTC loss \cite{graves2006connectionist} through fine-tuning.

The fine-tuned model can be used to transcribe speech into emotion-dependent units. For the purpose of speech emotion recognition, the recognized emotion-dependent units of an utterance are mapped to an emotion category by majority vote. For example, if most of the recognized units are of “happy”, then the utterance will be classified as “happy”.

\subsection{Phonetic units}

We tried four types of phonetic units for recognition: phonemes, broad phonetic classes, syllables, and the entire utterance. They are summarized in Figure 4 with examples.

\begin{figure}[t]
  \centering
  \includegraphics[width=\linewidth]{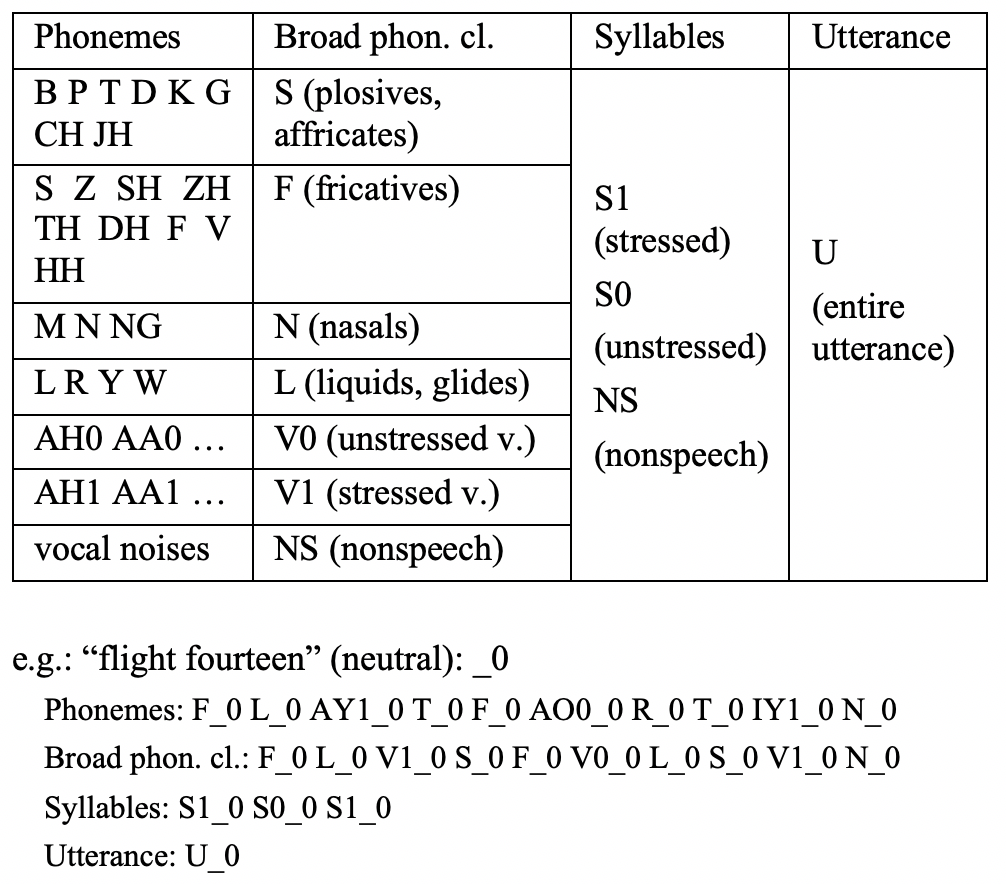}
  \caption{Phonetic units for speech emotion recognition.}
  \label{fig:speech_production}
\end{figure}

\section{Experiments on IEMOCAP}

\subsection{Data}
IEMOCAP is a benchmark emotion dataset in English. It consists of 12 hours of speech from 10 professional actors. Following the literature, we extracted 5531 utterances of four emotion types from the dataset: 1708 neutral, 1636 happy (also including excited), 1103 angry, and 1084 sad. 

\subsection{Fine-tuning}
The wav2vec 2.0 large model pre-trained on 960 hours of Librispeech audio, \textit{libri960\_big.pt}, was used for fine-tuning. The model was fine-tuned with 15k updates in all our experiments. For the first 10k updates only the output classifier is trained, after which the Transformer is also updated. The max\_tokens was set to 1m (which is equivalent to 62.5-second audio with sampling rate of 16 kHz), the learning rate was 5e-5.

\begin{figure}[t]
  \centering
  \includegraphics[width=\linewidth]{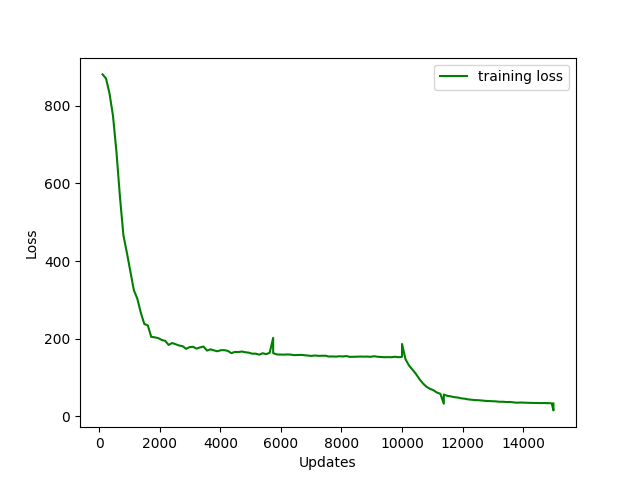}
  \caption{A typical training loss curve observed in fine-tuning.}
  \label{trainingloss}
\end{figure}

Figure~\ref{trainingloss} shows a typical training loss curve observed in fine-tuning.

To be consistent with the literature, we conducted 10-fold cross validation on the dataset. In each fold, the utterances of one speaker were used for testing, and the other nine speakers for fine-tuning.

\subsection{Results}
\subsubsection{Classification of emotion}
The performance of emotion recognition has been evaluated in terms of both weighted and unweighted accuracy. Weighted accuracy (WA) is the overall accuracy on the entire dataset while unweighted accuracy (UA) is the average of accuracies (recalls) of the emotion categories. 
The accuracies of models using different phonetic units are reported in Table~\ref{iemocapaccuracies}. 

\begin{table}[ht]
\caption{Emotion recognition accuracy from using different phonetic units.}
\begin{center}
\begin{tabular}{c|c|c}
\hline
\textbf{Phonetic units}&\textbf{WA}&\textbf{UA} \\
\hline
Phonemes&75.5\%&75.6\% \\
Broad phonetic classes &76.8\%&77.2\% \\
Syllables&75.4\%&75.4\% \\
Utterance&62.6\%&62.5\% \\
\hline
\end{tabular}
\end{center}
\label{iemocapaccuracies}
\end{table}

From Table~\ref{iemocapaccuracies} we can see that all models perform significantly better than most previously reported results, except the utterance model. In the utterance model, the entire utterance is a unit therefore there are only four emotion dependent units (one for each of the four emotions) for recognition. This is essentially a classification on the utterance without considering its phonetic content. The results demonstrate that phonetic units are helpful and should be incorporated in speech emotion recognition.

The best performance is from broad phonetic classes, and phonemes and syllables perform equally well. Table~\ref{confusionmatrix} is the confusion matrix of the emotion recognition results from fine-tuning broad phonetic class models. 

\begin{table}[t]
\caption{Confusion matrix of emotion recognition using broad phonetic classes.}
\label{confusionmatrix}
\begin{center}
\begingroup
\setlength{\tabcolsep}{1.5pt}
    \begin{tabular}{c|c c c c}
    \hline
    Target emotion & \multicolumn{4}{|c}{Predicted emotion} \\
     &   Neutral & Happy & Angry & Sad \\ \hline
   Neutral & \textbf{1189} & 298 & 100 & 121 \\
   Happy & 222 & \textbf{1328} & 64 & 22 \\
   Angry & 67 & 69 & \textbf{958} & 9 \\
   Sad & 195 & 93 & 25 & \textbf{771} \\ \hline
    \end{tabular}%
\endgroup
\end{center}
\end{table}

\subsubsection{Recognition of phonetic units}
Although Wav2vec 2.0 has achieved great success on speech recognition in terms of word error rate and phoneme error rate, it remains to be investigated whether the model can be fine-tuned to recognize other phonetic unites such as broad phonetic classes and syllables.

We evaluated the recognition results of phonemes, broad phonetic classes, and syllables using the NIST scoring toolkit SCTK \cite{SCTK}, both including and excluding emotions. The error rates are listed in Table~\ref{phoneerrors}.

\begin{table}[ht]
\caption{Recognition error rates of different phonetic units counting or not counting emotion. The percentages in brackets are substitution, deletion, and insertion error rates respectively.}
\begin{center}
\begin{tabular}{c|c|c}
\hline
\textbf{Phonetic units}&\textbf{with emotion}&\textbf{without emotion} \\
\hline
Phonemes&31.9\%&14.9\% \\
(N=194,473)&(24.2\%,5.0\%,2.8\%)&(6.6\%,5.2\%,3.0\%) \\
\hline
Broad phon. cl. &28.2\%&11.5\% \\
 (N=194,473)&(20.3\%,5.2\%,2.8\%)\%&(3.0\%,5.4\%,3.0\%) \\
\hline
Syllables&32.4\%&16.3\% \\
(N=78,943)&(21.4\%,8.8\%,2.2\%)& (4.9\%,9.0\%,2.4\%) \\
\hline
\end{tabular}
\end{center}
\label{phoneerrors}
\end{table}

If we don't count the emotion type of each recognized phonetic unit, the recognition error rates for phonemes, broad phonetic classes, and syllables are 14.9\%, 11.5\%, and 16.3\% respectively. The results show that Wav2vec 2.0 can be fine-tuned to recognize not only phonemes, but also coarser-grained or larger phonetic units such as broad phonetic classes and syllables, without a language model.

\section{Experiments on datasets in other languages}

\subsection{Data}

We re-trained models of phonemes, broad phonetic classes, and syllables on the entire dataset of IEMOCAP, with the same hyperparameters as used in the experiments above. We then tested these models on three other emotion datasets, EmoDB \cite{EmoDB}, KSUEmotions \cite{KSUEmotions}, and Mandarin Affective Speech \cite{MandarinAffective}.

EmoDB is a German dataset of emotional speech, containing utterances performed in seven target emotions (including neutral) by ten actors. KSUEmotions contains emotional Modern Standard Arabic (MSA) speech from 23 subjects, in six emotions. Mandarin Affective Speech contains recordings in five emotions by 68 Mandarin speakers. The datasets all contain neutral, happy, angry and sad emotions. The utterances in these emotions were extracted from these datasets for our experiments. The total number of utterances in each emotion and each dataset is listed in Table~\ref{datasize}.

\begin{table}[ht]
\caption{The number of utterances in each emotion and each dataset.}
\begin{center}
\begin{tabular}{c|c|c|c|c}
\hline
\textbf{Datasets}&\textbf{Neutral}&\textbf{Happy}&\textbf{Angry}&\textbf{Sad} \\
\hline
EmoDB (German) &79&71&127&62 \\
KSU (Arabic) &280&280&280&280 \\
Affective (Mandarin) &4079&4080&4080&4080 \\
\hline
\end{tabular}
\end{center}
\label{datasize}
\end{table}

\subsection{Results}

Table~\ref{crosslingualresults} lists the recognition results on the three datasets. 

\begin{table}[t]
\caption{The (weighted) accuracies of emotion recognition on three other datasets using models trained on IEMOCAP (English).}
\label{crosslingualresults}
\begin{center}
\begingroup
\setlength{\tabcolsep}{1.5pt}
    \begin{tabular}{c|c c c}
    \hline
     Datasets & \multicolumn{3}{|c}{IEMOCAP models} \\
     &  Phonemes & Broad & Syllables \\ \hline
   EmoDB (German) & 60.4\%  & 66.7\% & 68.4\% \\
   KSU (Arabic) & 54.2\% & 60.6\% & 55.0\% \\
   Affective (Mandarin) & 40.7\% & 39.9\% & 41.0\% \\
   \hline
    \end{tabular}%
\endgroup
\end{center}
\end{table}

The Wav2vec 2.0 pre-trained models were trained on audio with sampling rate of 16 kHz. Recordings in the IEMOCAP, EmoDB, and KSUEmotions are also sampled at 16 kHz. The Mandarin Affective Speech dataset, however, uses 8 kHz sampling rate. We upsampled the audio to 16 kHz for testing this dataset. Although the sampling rate is the same, the audio upsampled from 8 kHz to 16 kHz contains no components in frequencies between 4 and 8 kHz, which is dramatically different from the training data. This may explain why the IEMOCAP models performed poorly on the Mandarin dataset. We also directly tested on 8k Hz audio with 16 kHz models. However, the results were even worse. 

We conducted a 7-fold cross validation experiment on the Mandarin Affective Speech dataset, by partitioning the dataset into training and test sets. We used initials and finals as the phonetic units in the experiment. The overall (weighted) accuracy was 75.2\%. This result shows that the poor performance of IEMOCAP models on this dataset is not due to the nature of the dataset. The mismatch of sampling rate is probably the main factor responsible for the poor performance, together with other possible factors such as linguistic and cultural differences.

The models performed reasonably well on the German and Arabic datasets, suggesting that our method has the potential for cross-lingual emotion recognition. The models of broad phonetic classes and syllables seem to work better than the phoneme models. This may be because broad phonetic classes and syllables are language-general whereas phonemes are language-specific. It may also be because there are more training data for broad phonetic classes and syllables than phonemes, therefore the models of broad phonetic classes and syllables are more robust. Further studies are needed to test these hytpotheses.

\section{Conclusions and discussion}
We propose a method for emotion recognition through fine-tuning wav2vec 2.0 for recognition of emotion-dependent phonetic units, including phonemes, broad phonetic classes, syllables, as well as the entire utterance. The models of phonemes, broad phonetic classes, and syllables all significantly outperform the utterance model, demonstrating that phonetic units are helpful and should be incorporated in speech emotion recognition.

The best performance is from broad phonetic classes. The advantages of using broad phonetic classes for other tasks such as speaking rate estimation \cite{yuan&liberman} and speech enhancement \cite{enhancement} have been reported in the literature. Using broad phonetic classes may force the model to disregard differences of phonemes within a broad phonetic class, but also encourage the model to pay attention to the phonetic variability among broad phonetic classes in emotion recognition. It will be interesting to further investigate the optimal set of broad phonetic classes for the task of emotion recognition. 

Models of broad phonetic classes and syllables outperform phonemes in the setting of cross-lingual emotion recognition. This may be because broad phonetic classes and syllables are language-general whereas phonemes are language-specific, or because more training data are available for broad phonetic classes and syllables than phonemes. Further research is needed to test these hypotheses.

Our study also shows that Wav2vec 2.0 can be fine-tuned to recognize not only phonemes, but also coarser-grained or larger phonetic units such as broad phonetic classes and syllables, without a language model.

The proposed method of fine-tuning Wav2vec 2.0 for emotion recognition achieved a significant improvement over most previously reported results on IEMOCAP. In the future we will investigate how to make fine-tuned models more robust across datasets where there are mismatches in recording conditions as well as sampling rate.

\bibliographystyle{IEEEtran}

\bibliography{mybib}

\begin{thebibliography}{10}
\providecommand{\url}[1]{#1}
\csname url@samestyle\endcsname
\providecommand{\newblock}{\relax}
\providecommand{\bibinfo}[2]{#2}
\providecommand{\BIBentrySTDinterwordspacing}{\spaceskip=0pt\relax}
\providecommand{\BIBentryALTinterwordstretchfactor}{4}
\providecommand{\BIBentryALTinterwordspacing}{\spaceskip=\fontdimen2\font plus
\BIBentryALTinterwordstretchfactor\fontdimen3\font minus
  \fontdimen4\font\relax}
\providecommand{\BIBforeignlanguage}[2]{{%
\expandafter\ifx\csname l@#1\endcsname\relax
\typeout{** WARNING: IEEEtran.bst: No hyphenation pattern has been}%
\typeout{** loaded for the language `#1'. Using the pattern for}%
\typeout{** the default language instead.}%
\else
\language=\csname l@#1\endcsname
\fi
#2}}
\providecommand{\BIBdecl}{\relax}
\BIBdecl

\bibitem{Ververidis&Kotropoulos}
D.~Ververidis and C.~Kotropoulos, ``Emotional speech recognition: Resources,
  features, and methods,'' \emph{Speech Communication}, vol.~48, pp.
  1162--1181, 09 2006.

\bibitem{Ayadietal}
M.~Ayadi, M.~S. Kamel, and F.~Karray, ``Survey on speech emotion recognition:
  Features, classification schemes, and databases,'' \emph{Pattern
  Recognition}, vol.~44, pp. 572--587, 03 2011.

\bibitem{Koolagudi&Rao}
S.~Koolagudi and K.~Rao, ``Emotion recognition from speech: A review,''
  \emph{International Journal of Speech Technology}, vol.~15, pp. 90--117, 06
  2012.

\bibitem{Anagnostopoulosetal}
C.-N. Anagnostopoulos, T.~Iliou, and I.~Giannoukos, ``Features and classifiers
  for emotion recognition from speech: a survey from 2000 to 2011,''
  \emph{Artificial Intelligence Review}, vol.~43, pp. 155--177, 02 2012.

\bibitem{akccay2020speech}
M.~B. Ak{\c{c}}ay and K.~O{\u{g}}uz, ``Speech emotion recognition: Emotional
  models, databases, features, preprocessing methods, supporting modalities,
  and classifiers,'' \emph{Speech Communication}, vol. 116, pp. 56--76, 2020.

\bibitem{Eybenetal}
F.~Eyben, M.~Wöllmer, and B.~Schuller, ``opensmile -- the munich versatile and
  fast open-source audio feature extractor,'' in \emph{MM'10 - Proceedings of
  the ACM Multimedia 2010 International Conference}, 01 2010, pp. 1459--1462.

\bibitem{busso2008iemocap}
C.~Busso, M.~Bulut, C.-C. Lee, A.~Kazemzadeh, E.~Mower, S.~Kim, J.~N. Chang,
  S.~Lee, and S.~S. Narayanan, ``Iemocap: Interactive emotional dyadic motion
  capture database,'' \emph{Language resources and evaluation}, vol.~42, no.~4,
  pp. 335--359, 2008.

\bibitem{wangetal}
J.~Wang, M.~Xue, R.~Culhane, E.~Diao, J.~Ding, and V.~Tarokh, ``Speech emotion
  recognition with dual-sequence lstm architecture,'' in \emph{ICASSP 2020}, 05
  2020, pp. 6474--6478.

\bibitem{liuetal}
J.~Liu, Z.~Liu, L.~Wang, L.~Guo, and J.~Dang, ``Speech emotion recognition with
  local-global aware deep representation learning,'' in \emph{ICASSP 2020}, 05
  2020, pp. 7174--7178.

\bibitem{luetal}
Z.~Lu, L.~Cao, Y.~Zhang, C.-C. Chiu, and J.~Fan, ``Speech sentiment analysis
  via pre-trained features from end-to-end asr models,'' in \emph{ICASSP 2020},
  05 2020, pp. 7149--7153.

\bibitem{Pappagarietal}
R.~Pappagari, T.~Wang, J.~Villalba, N.~Chen, and N.~Dehak, ``X-vectors meet
  emotions: A study on dependencies between emotion and speaker recognition,''
  in \emph{ICASSP 2020}, 05 2020, pp. 7169--7173.

\bibitem{Yehetal}
S.~L. Yeh, Y.-S. Lin, and C.-C. Lee, ``A dialogical emotion decoder for speech
  motion recognition in spoken dialog,'' in \emph{ICASSP 2020}, 05 2020, pp.
  6479--6483.

\bibitem{Shahetal}
M.~Shah, M.~Tu, V.~Berisha, C.~Chakrabarti, and A.~Spanias, ``Articulation
  constrained learning with application to speech emotion recognition,''
  \emph{EURASIP Journal on Audio, Speech, and Music Processing}, vol. 2019, pp.
  1--17, 08 2019.

\bibitem{Bitouketal}
D.~Bitouk, R.~Verma, and A.~Nenkova, ``Class-level spectral features for
  emotion recognition,'' \emph{Speech communication}, vol.~52, pp. 613--625, 07
  2010.

\bibitem{Schulleretal}
B.~Schuller, B.~Vlasenko, D.~Arsic, G.~Rigoll, and A.~Wendemuth, ``Combining
  speech recognition and acoustic word emotion models for robust
  text-independent emotion recognition,'' in \emph{2008 IEEE International
  Conference on Multimedia and Expo}, 05 2008, pp. 1333 -- 1336.

\bibitem{Dhamyal2020ThePB}
H.~Dhamyal, S.~A. Memon, B.~Raj, and R.~Singh, ``The phonetic bases of vocal
  expressed emotion: natural versus acted,'' \emph{ArXiv}, vol. abs/1911.05733,
  2020.

\bibitem{Vlasenko&Wendemuth}
B.~Vlasenko and A.~Wendemuth, ``Determining the smallest emotional unit for
  level of arousal classification,'' in \emph{2013 Humaine Association
  Conference on Affective Computing and Intelligent Interaction}, 09 2013, pp.
  734--739.

\bibitem{baevski2020wav2vec}
A.~Baevski, H.~Zhou, A.~Mohamed, and M.~Auli, ``wav2vec 2.0: A framework for
  self-supervised learning of speech representations,'' \emph{arXiv preprint
  arXiv:2006.11477}, 2020.

\bibitem{panayotov2015librispeech}
V.~Panayotov, G.~Chen, D.~Povey, and S.~Khudanpur, ``Librispeech: an asr corpus
  based on public domain audio books,'' in \emph{2015 IEEE international
  conference on acoustics, speech and signal processing (ICASSP)}.\hskip 1em
  plus 0.5em minus 0.4em\relax IEEE, 2015, pp. 5206--5210.

\bibitem{graves2006connectionist}
A.~Graves, S.~Fern{\'a}ndez, F.~Gomez, and J.~Schmidhuber, ``Connectionist
  temporal classification: labelling unsegmented sequence data with recurrent
  neural networks,'' in \emph{Proceedings of the 23rd international conference
  on Machine learning}, 2006, pp. 369--376.

\bibitem{SCTK}
``The nist scoring toolkit,'' \url{https://github.com/usnistgov/SCTK}.

\bibitem{EmoDB}
F.~Burkhardt, A.~Paeschke, M.~Rolfes, W.~Sendlmeier, and B.~Weiss, ``A database
  of german emotional speech,'' in \emph{9th European Conference on Speech
  Communication and Technology}, vol.~5, 01 2005, pp. 1517--1520.

\bibitem{KSUEmotions}
A.~H. Meftah, Y.~A. Alotaibi, and S.-A. Selouani, ``Ksuemotions,''
  \url{https://catalog.ldc.upenn.edu/LDC2017S12}.

\bibitem{MandarinAffective}
Y.~Yang, Z.~Wu, T.~Wu, and D.~Li, ``Mandarin affective speech,''
  \url{https://catalog.ldc.upenn.edu/LDC2007S09}.

\bibitem{yuan&liberman}
J.~Yuan and M.~Liberman, ``Robust speaking rate estimation using broad phonetic
  class recognition,'' in \emph{ICASSP 2010}, 04 2010, pp. 4222 -- 4225.

\bibitem{enhancement}
Y.-J. Lu, C.-F. Liao, X.~lu, J.-w. Hung, and Y.~Tsao, ``Incorporating broad
  phonetic information for speech enhancement,'' \emph{ArXiv}, 08 2020.

\end{thebibliography}


\end{document}